\documentclass[accepted]{article}

\usepackage{microtype}
\usepackage{graphicx}
\usepackage{subfigure}
\usepackage{booktabs} 

\usepackage{hyperref}
\usepackage{enumitem}

\usepackage{icml2022}

\usepackage{amsmath}
\usepackage{amssymb}
\usepackage{mathtools}
\usepackage{amsthm}

\usepackage[capitalize,noabbrev]{cleveref}

\theoremstyle{plain}

\theoremstyle{definition}

\theoremstyle{remark}

\usepackage[textsize=tiny]{todonotes}
\usepackage{multirow}
\usepackage{float}
\usepackage{stfloats}
\icmltitlerunning{Data-Efficient Protein 3D Geometric Pretraining via Refinement of Diffused Protein Structure Decoy}

\begin{document}

\twocolumn[
\icmltitle{Data-Efficient Protein 3D Geometric Pretraining via Refinement of \\ Diffused Protein Structure Decoy}



\icmlsetsymbol{equal}{*}

\begin{icmlauthorlist}
\icmlauthor{Yufei Huang}{equal,ZJU,westlake}
\icmlauthor{Lirong Wu}{equal,ZJU,westlake}
\icmlauthor{Haitao Lin}{ZJU,westlake}
\icmlauthor{Jiangbin Zheng}{ZJU,westlake}
\icmlauthor{Ge Wang}{ZJU,westlake}
\icmlauthor{Stan Z. Li}{westlake}
\end{icmlauthorlist}

\icmlaffiliation{ZJU}{Department of Computer Science, Zhejiang University, Hangzhou, China}
\icmlaffiliation{westlake}{AI Research and Innovation Lab, Westlake University, Hangzhou, China}

\icmlcorrespondingauthor{Stan Z. Li}{Stan.ZQ.Li@westlake.edu.cn}

\icmlkeywords{Protein Representation learning, Structure Pretrain task, Protein Structure Refinement}

\vskip 0.3in
]



\printAffiliationsAndNotice{\icmlEqualContribution}  

\begin{abstract}
Learning meaningful protein representation is important for a variety of biological downstream tasks such as structure-based drug design. Having witnessed the success of protein sequence pretraining, pretraining for structural data which is more informative has become a promising research topic. However, there are three major challenges facing protein structure pretraining: insufficient sample diversity, physically unrealistic modeling, and the lack of protein-specific pretext tasks. To try to address these challenges, we present the 3D Geometric Pretraining. In this paper, we propose a unified framework for protein pretraining and a \emph{3D geometric-based}, \emph{data-efficient}, and \emph{protein-specific} pretext task: \emph{RefineDiff} (Refine the Diffused Protein Structure Decoy). After pretraining our geometric-aware model with this task on limited data(less than 1\% of SOTA models), we obtained informative protein representations that can achieve comparable performance for various downstream tasks.
\end{abstract}

\section{Introduction}
\label{Introduction}

Proteins are involved in various important life processes, such as immune response and DNA replication, so understanding proteins is important for deciphering the mystery of life and treating various diseases.\cite{jumper2021highly} With the advancement of deep learning technology and the accumulation of protein data, the use of deep learning to explore the unknown protein space has gained increasing attention, with one of the keys being the acquisition of information-rich protein representations~\cite{jumper2021highly,meier2021language,rao2021msa}. Protein sequence pretraining has been a huge success in the last few years. Researchers have migrated the powerful techniques of sequence pretraining in natural language processing to protein sequence modeling, pretraining amazing protein sequence representation models with increasingly large data (up to hundreds of millions) and increasingly large models (as large as 15B parameters)\cite{rao2020transformer,rao2021msa,lin2022language}. Whereas protein structure determines function, Protein structure data contains more information than simple sequence data, and inspired by the success of sequence pretraining, structure pretraining became a natural idea. The emergence of high-precision structure prediction tools in recent years has enabled researchers to obtain large amounts of new structural data with reasonable confidence, which has led the idea of structure pretraining to reality.\cite{zhang2022protein,chen2022structure,cheng2021graphms} 

Pretraining usually consists of three elements: data, model, and pretext task. Protein features or properties can be broadly classified as scalars (e.g., amino acid type, angle, the distance between amino acids) and vectors (e.g., coordinates of protein atoms, the orientation of amino acids, etc.). Rethinking the development of protein pretraining in conjunction with the above two aspects, we find that:\\ 
\emph{\textbf{(1) Phase \uppercase\expandafter{\romannumeral1}: Big Data, model and pretext task are in 1D}}: in the beginning, protein sequence pretraining models were trained using huge amounts of data (hundreds of millions of data), based on a non-geometry-aware Transformer, using scalar properties prediction on protein sequences as pretext task (i.e., tasks such as amino acid type prediction)\cite{rives2019biological,rao2020transformer,rao2021msa,lin2022language}.\\ 
\emph{\textbf{(2) Phase \uppercase\expandafter{\romannumeral2}: Much less data, model and pretext task are in 2D}}: Subsequent protein structure pretraining models are trained using much less data (millions or so), based on a 2D Graph Neural Network(GNN), using scalar attribute prediction on protein structures as pretext task (i.e., tasks such as distance prediction, dihedral angle prediction, etc.)\cite{zhang2022protein,hermosilla2022contrastive,cheng2021graphms}. 

Currently, 2D protein structure pretraining has made considerable progress but still faces three major problems.\\ \emph{\textbf{(1) Data Homogeneity}}: The gains from expanding the pretraining dataset can rapidly decline while the costs increase significantly. This is partly due to the severe lack of data diversity in large structure datasets when modeling in 2D.\cite{zhang2022protein,hsu2022learning} \emph{\textbf{(2) Physically unrealistic modeling}}: For example, in Distance Map prediction, the distance given by the model may violate the triangle inequality\cite{jumper2021highly,zhou2022uni,lin2022diffbp} which indicates a lack of inductive bias. \emph{\textbf{(3) limited pretext tasks}}: The 2D geometric pretraining constrains the pretext task to the framework of 2D GNN pretraining while not being able to explore protein-specific tasks(typically in 3D) with structure biology insights\cite{wu2022survey}.

\emph{\textbf{(3) Phase \uppercase\expandafter{\romannumeral3}: Towards pretraining in 3D, costly to train yet data-efficient}}: To solve the problems above, pretraining in 3D is a promising direction although there have been few attempts in 3D geometric pretraining. Meanwhile, 3D Geometric representation models\cite{jing2020learning,hsu2022learning,jumper2021highly} of proteins and 3D Geometric pretraining in molecules\cite{zhou2022uni} are developing rapidly recently. Considering the current dilemmas, the trends in the field, and the models available, we propose 3D Geometric Pretraining on proteins.

In this paper, we propose a \emph{data-efficient}, \emph{3D geometric-based}, and \emph{protein-specific} pretext task named \texttt{RefineDiff} \emph{(Refine the Diffused protein Decoy)}. Due to the limitation of computational resources, we pretrain an orientation-aware roto-translation invariant transformer\cite{jumper2021highly} on limited data (1\% of SOTA structure pretrain models) using this pretraining task, but it achieved comparable results on various downstream tasks. Our main contributions can be summarized as follows:

\vspace{-1em}                            
\begin{itemize}[itemsep=2pt,topsep=0pt,parsep=0pt,partopsep=1em]
\item We present a unified framework for protein pre-training and representation learning that provides a clear overview of previous work and reveals promising future directions.
\item We propose a \emph{novel}, \emph{data-efficient}, and \emph{protein-specific} pretext task for 3D Geometric pretraining.
\item We pretrained an orientation-aware roto-translation invariant transformer on limited data, and systematically tested and compared its performance on various downstream tasks. The results initially reveal the power of 3D Geometric pretraining.
\end{itemize}

\vspace{-1.5em}
\section{Related Work}
\vspace{-0.3em}
\noindent \textbf{Protein Structure Refinement.}
Protein Structure Refinement(PSR) is a long-standing problem in computational structural biology. This task inspired us to propose \emph{RefineDiff} as a protein 3D geometric pretraining task. The PSR Challenge is to increase the accuracy of the predicted protein structure or known as \emph{Decoy}\cite{hiranuma2021improved}. In the past, PSR methods were mainly based on molecular dynamics simulations as well as conformational sampling, which tried to approximate the true folding energy landscape in terms of empirical force fields and statistics\cite{adiyaman2019methods}. Subsequent deep learning methods have attempted to directly approximate the energy landscape\cite{roney2022state}. Before Alphafold2, deep learning was typically combined with traditional methods to model at the pseudo-3D level. Using neural networks to guide or directly optimize the protein scalar feature Distance Map, on top of which tools such as Rosetta were used to sample lower energy conformations\cite{hiranuma2021improved,jing2020learning}. Following Alphafold2, the field shifted to direct modelling in 3D, using equivariant neural networks to directly obtain atomic-level optimization results\cite{wu2022atomic}.

\noindent \textbf{DDPM for Protein Structure.}
Denoising Diffusion Probability Model (DDPM)\cite{sohl2015deep,ho2020denoising}, a recently developed generative model, has been applied to protein 3D structure generation. Its forward diffusion process inspired \emph{RefineDiff}. Unusually, the first pieces of work are the direct generation of 3D structures(not pseudo-3D features)\cite{anand2022protein}, and until now, it's been the 3D generation that dominated. Subsequent work has seen further developments in application areas, forward diffusion processes\cite{luo2022antigen}, sampling\cite{trippe2022diffusion}, conditional generation\cite{ingraham2022illuminating}, network structures\cite{watson2022broadly}, etc. Efficient DDPM based on pseudo-3D features (like distance maps\cite{lee2022proteinsgm}, and torsion angles\cite{wu2022protein}) have also emerged during this period. Recently, RFDiffusion\cite{watson2022broadly} reveals the connection between denoising models and structure prediction models, and Chroma\cite{ingraham2022illuminating} significantly developed the protein conditional generation model.

\noindent \textbf{Protein Pretraining.}
With the development of protein representation models, pretraining was naturally introduced into the field. The first great successes were protein sequence pretraining. The Evolutionary Scale Modeling (ESM) family of models has achieved remarkable success in difficult downstream tasks such as mutation prediction\cite{meier2021language} and protein structure prediction models\cite{hsu2022learning}. With the development of structure prediction tools as well as GNN pretraining, structure pretraining has taken off, and several recent works have advanced the development of pseudo-3D structure pretraining. They typically downscale 3D structures into scalar features on 2D GNNs and migrate the pretext task of 2D GNN Pretraining\cite{zhang2022protein,hermosilla2022contrastive,chen2022structure}. Thanks to the development of geometric deep learning, end-to-end 3D pretraining started a limited exploration\cite{mansoor2021toward}, facing the difficulty of insufficient 3D pretraining tasks. The 3D molecular pretraining model Uni-Mol\cite{zhou2022uni} has shown powerful representation capabilities, and we believe that protein 3D pretraining is an equally promising direction.
\begin{figure*}[!tbp]
    \centering
    \includegraphics[width=\linewidth]{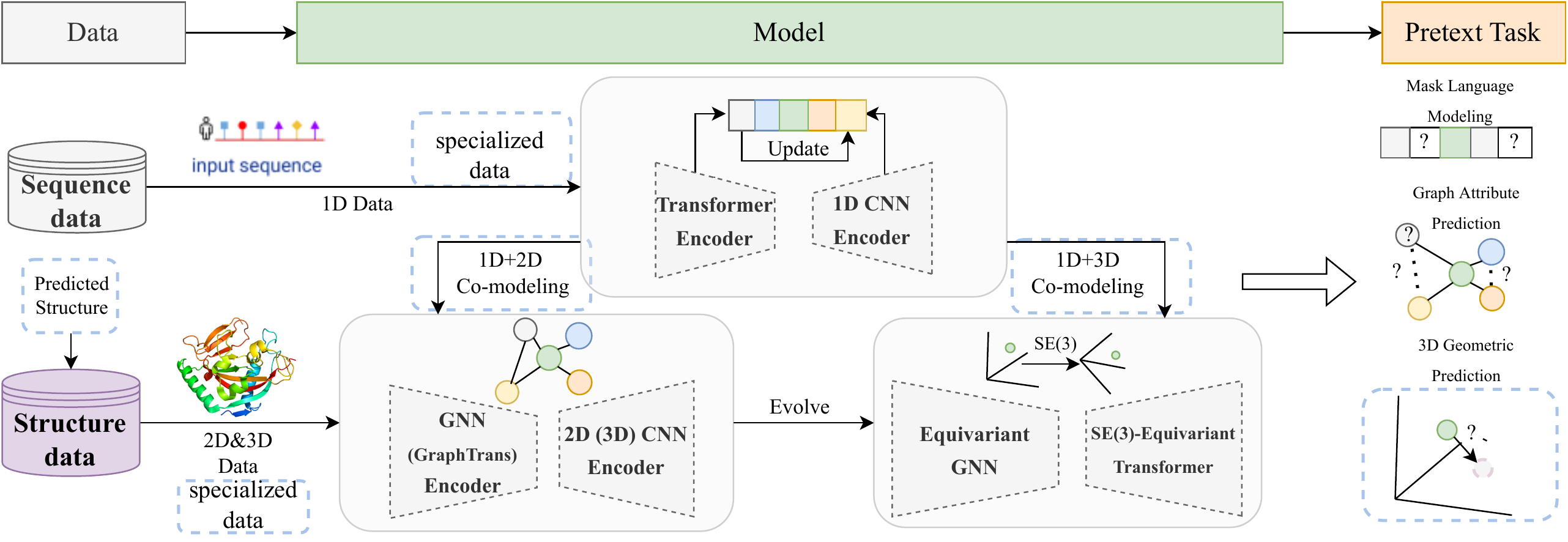}
    \vspace{-1em}
    \caption{A unified framework for protein pretraining. The part framed by the blue line is the direction worth noting. 1D means protein sequence, 2D means 2D protein graph which is based on KNN, 3D means modeling in 3D space or \textbf{3D geometric aware}}
    \vspace{-1em}
    \label{fig:1}
\end{figure*}
\vspace{-0.8em}
\section{A Unified Framework for Protein Pretraining}
\vspace{-0.3em}
\subsection{Notations and Problem Statement}
Protein data can be modeled at multiple levels: sequence, amino acid level, full atom level, etc. Here we model proteins uniformly as an Attributed Relational Graph: $\mathcal{G} = (\mathcal{V}, \mathcal{E}, \mathcal{N}, \mathcal{R})$, where $\mathcal{V}$ represents the ordered set of graph nodes (can be amino acids or atoms) and $\mathcal{E} \in \mathcal{V} \times \mathcal{V}$ represents the corresponding set of edges connecting the nodes (some relationship between nodes, e.g., distance less than 4 \AA). every vertex $v\in\mathcal{V}$ in $\mathcal{G}$ can have both scalar and vector attributes $\mathbf{n}_v=(S_v, V_v) \in \mathcal{N}$, where $S_v \in \mathbb{R}^S$ and $V_v \in \mathbb{R}^{3\times V}$. Similarly, each edge $e \in \mathcal{E}$ have attributes $\mathbf{r}_v=(S_e, V_e) \in \mathcal{R}$, where $S_e \in \mathbb{R}^N$ and $V_e \in \mathbb{R}^{3\times T}$. $\mathcal{G}$ can contain empty sets. When the sets $\mathcal{E}$ and $\mathcal{R}$ are empty sets, $\mathcal{G}$ degenerates to a single sequence representation. Furthermore, if $\mathcal{N}$ contains only amino acid composition, $\mathcal{G}$ degenerates to the amino acid sequence.

Protein Pretraining usually consists of three elements: data $\mathcal{D}$, representation model $f_{\theta}(\cdot)$, and pretext tasks as losses

$\{\mathcal{L}_{pre}^{\uppercase\expandafter{\romannumeral1}}(\theta, \mathcal{D}), \mathcal{L}_{pre}^{\uppercase\expandafter{\romannumeral2}}(\theta, \mathcal{D}), \cdots, \mathcal{L}_{pre}^{\mathcal{T}}(\theta, \mathcal{D})\}$. Each task corresponds to a specific projection head $\{g_{T}(\cdot)\}_{T=1}^{\mathcal{T}}$. Protein pretraining is generally performed in two steps: (1) Pretraining the representation model $f_\theta(\cdot)$ with (a) pretext task(s) in pretraining dataset $D_{pre}$; and (2) Fine-tuning the pretrained representation model $f_{\theta_{pre}}(\cdot)$ with a prediction head $g_{downstream}(\cdot)$ under the supervision of a specific downstream task $\mathcal{L}_{task}(\theta, \mathcal{D}_{task})$. The whole process can be formulated as
\setlength\abovedisplayskip{0.3em}
\setlength\belowdisplayskip{0.3em}
\begin{equation}
\begin{small}
\begin{aligned}
\theta^{*} = & \arg \min _{\theta} \mathcal{L}_{task}(\theta_{pre}, \mathcal{D}_{task}), \\ \text{s.t.} \text{ } \text{ } \theta_{pre} = &\mathop{\arg\min}_{\theta} \sum_{k=1}^K \lambda_k \mathcal{L}_{pre}^{(k)}(\theta, \mathcal{D})
\label{equ:1}
\end{aligned}
\end{small}
\end{equation}
where $\{\lambda_k\}_{k=1}^K$ are task weight hyperparameters. In particular, if we set $K = 1$ and $\mathcal{L}_{pre}^{(1)}(\theta, \mathcal{D}) = \mathcal{L}_{task}(\theta, \mathcal{D}_{task})$, it is equivalent to train from scratch on a downstream task like many previous works on protein representation, which can be considered a special case of our framework.

\subsection{Equivariant and Invariant}
The \emph{scalar} and \emph{vector} attributes we mentioned before should be strictly defined as \emph{invariant} or \emph{equivariant} attribute under rotation or translation of the protein. Formally, $f: \mathbb{R}^3 \rightarrow \mathbb{R}^S$ is an \emph{invariant} function \textit{w.r.t} SE(3) group, if for any rotation and translation transformation in the group, which is represented by  $\mathbf{\mathrm{R}}$ as orthogonal matrices and $\mathbf{\mathrm{t}} \in\mathbb{R}^3$   respectively, $f(\mathbf{\mathrm{R}} \mathbf{x} + \mathbf{\mathrm{t}}) = f(\mathbf{x})$. Attributes generated by \emph{invariant} function are \emph{invariant} and \emph{scalar}. If $f: \mathbb{R}^3 \rightarrow \mathbb{R}^3$ is an \emph{equivariant} \textit{w.r.t.} SE(3) group, then $f(\mathbf{\mathrm{R}} \mathbf{x} + \mathbf{\mathrm{t}}) = \mathbf{\mathrm{R}} f(\mathbf{x}) + \mathbf{\mathrm{t}}$. Attributes generated by \emph{equivariant} function are \emph{equivariant} and \emph{vector}. $f(\cdot)$ can be neural networks, and definitions still apply.
\subsection{A Unified Framework for Protein Pretraining}
\label{framework}
Although protein pretraining is growing fast, it is growing wildly, unlike NLP and graph representation learning, where there are relatively clear frameworks and directions for development. Therefore, we hope to integrate previous works into a clear framework, rethink their essential contributions, and hopefully provide valuable suggestions for the further development of the field.
Here, We present a unified framework for protein pretraining, considering data, model architecture, and pretext tasks.

\vspace{-0.3em}
\textbf{Data.}
After the boom of sequence pretraining, protein pretraining is shifting from sequence-based to structure-based. However, there is still a gap in the quantity and quality of structural data compared to sequence data, so there are also explorations to combine sequence and structural information. Meanwhile, pretraining is gradually evolving from generic to specialized, such as antibody sequence pretraining\cite{ruffolo2021deciphering} and protein pocket structure pretraining\cite{zhou2022uni}.

\noindent \textbf{Model Architecture.}
The model architecture has changed with the data and has evolved from sequence transformer, GNN, to ENN, and then to co-modeling. Co-modeling currently focuses on a) initializing sequence pretraining model embedding as structural model features; b) modeling sequence and structure with different networks (basically, sequence representation as structure node feature or pair representation as sequence attention bias).

\noindent \textbf{Pretext Task.}
The development of pretext task has also witnessed several stages: a) 1D: Mask Language Modeling and its variants, supervised task based on evolutionary information such as multiple sequence matching and protein families b) 2D: Migrated 2D GNN pretraining task (node attribute prediction, edge attribute prediction), pretraining task based on contrast learning c) 1D+2D: Mask Inverse Folding d) 3D: Coordinate Denoising.

\noindent \textbf{Future Direction.}
Sequence-based pretraining is maturing, but structure-based pretraining is just beginning. Promising future directions include a) domain-specific pretraining; b) methods to better exploit large-scale predictive structure data; c) efficient neural networks; d) novel architecture on 3D graph data; e) 3D pretext task design; f) co-modeling pretraining (via network architecture, especially via pretext tasks). These advances may be related, and many have substantial implications for general deep learning. Based on this framework, we discovered that while data and models are available for 3D Geometric Pretraining, an appropriate pretext task is absent. Therefore, we propose \emph{RefineDiff}.

\vspace{-1em}
\begin{figure}[!htbp]
	\begin{center}
		\includegraphics[width=0.5\linewidth]{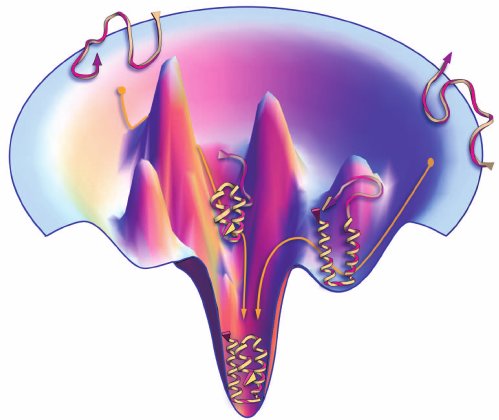}
	\end{center}
	\vspace{-1em}
	\caption{A illustration of protein folding energy landscape\cite{dill2012protein}}
	\vspace{-0.5em}
	\label{fig:3}
\end{figure}
\begin{figure*}[!tbp]
    \centering
    \includegraphics[width=\linewidth]{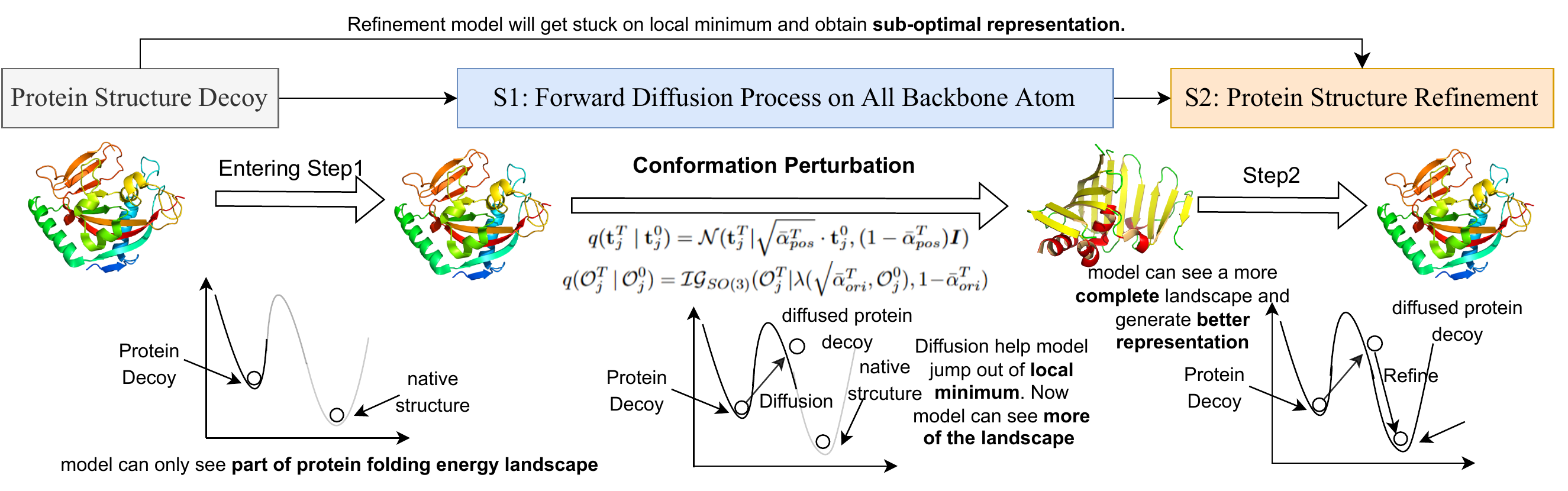}
    \vspace{-1em}
    \caption{The motivation and general process for RefineDiff.}
    \vspace{-1em}
    \label{fig:2}
\end{figure*}
\vspace{-0.8em}

\vspace{-0.7em}
\section{Methodology}
\vspace{-0.3em}

\subsection{Motivation behind RefineDiff}
One of the primary benefits of 3D Geometric Pretraining is \textbf{physical-level realistic modeling} that can learn from the original dimension of the data without losing information due to dimensionality reduction. Furthermore, it enables us to think about physically meaningful pretext tasks and learn generic representations while solving real-world physical challenges. So we started to consider the physics-based tasks of protein structural biology, and we found Protein Structure Refinement (PSR). The challenge of PSR is to estimate the folding energy landscape and the direction of energy minimization (expressed via coordinates shift), which is also the source and direction of the folding dynamics. 

PSR naturally becomes a potential choice for the pretext task. We hope that the model will learn information about the protein folding energy landscape through this task and thus obtain a more descriptive representation of the protein folding state for a variety of downstream tasks. However, the direct use of PSR as the pretext task tends to make the representation model quickly fall into the \textbf{local optimum}. We found in the pretraining process that when the accuracy of the predicted structure reached a certain level, it was difficult for the model to optimize it further; continuing pretraining at this point also resulted in little or no improvement in performance on downstream tasks \ref{ablation study}.

The main reason for this is that the protein decoys with reasonable accuracy are positioned at a local minimum of the protein folding landscape, and the neural network struggles to overcome the energy barrier to find conformations with lower energies. As a result, the representation model fails to comprehend a complete energy landscape, and the representation of the protein folding state is unable to become more accurate, failing to produce a more meaningful representation of the protein structure. 

Taking a step deeper, we can further understand this optimal representation problem from the perspective of information bottleneck theory following \cite{dubois2020learning}. Information Bottleneck (IB) is an information-theoretic method that explains deep representation learning under supervision. IB is based on the idea that, in order to prevent potential over-fitting, a representation Z should be as highly informative about Y as possible (sufficient) while including no additional information about X (minimum). The sets of sufficient representations $\mathcal{S}$ and minimal sufficient representations $\mathcal{M}$ are defined as follows:
\vspace{0.5em}
\begin{equation}
\begin{small}
\begin{aligned}
 \mathcal{S} := \arg\max _{Z^{\prime}} I[Y;Z^{\prime}] \ \ \text{and,} \\ \mathcal{M} := \arg\min _{Z^{\prime}\in\mathcal{S}} I[X;Z^{\prime}]
\label{equ:2}
\end{aligned}
\end{small}
\end{equation}
To obtain optimal representation, the IB criterion which representation models learn to minimize can be formulated as the Lagrangian relaxation of Eq.\ref{equ:2}:
\vspace{0.5em}
\begin{equation}
\begin{small}
\mathcal{L}_{IB} := -I[Y;Z] + \beta * I[X;Z]
\label{equ:3}
\end{small}
\end{equation}

In the setting of PSR, mutual information $I[Y;Z]$ corresponds to the similarity $S[Y;D]$ where Z is the decoded protein structure decoy from representation Z using the best possible structure decoder. So we can rewrite the Eq.\ref{equ:3} as:
\vspace{0.5em}
\begin{equation}
\begin{small}
\mathcal{L}_{IB} := -S[Y;D] + \beta * S[X;D]
\label{equ:4}
\end{small}
\end{equation}

When the input X is structurally similar to the label Y (i.e., structure decoys have acceptable accuracy), $S[Y;D]$ and $S[X;D]$ will be more likely to have the same monotonicity with respect to Z, leading $\mathcal{L}_{IB}$ into a dilemma of difficult trade-offs which block the learning of structure representation model.

To solve the above problem of PSR, we perform a forward diffusion process on the protein structure decoys to help them jump out of the local optimum before the structure refinement step, and finally propose a novel 3D Geometric protein pretext task named \emph{RefineDiff}: Refine Diffused Structure Decoy. 

The forward diffusion process of protein structure decoys can help them escape from the local minima of the folding energy landscape in which they were initially located and assist the model in finding the following local minima with lower energy, i.e., possibly closer to the conformation of the real structure. In this process, the model is able to see a complete energy landscape and obtain a more accurate representation of the folding state, resulting in a more generalized and informative representation. 

Additionally, from the viewpoint of information bottleneck theory, the forward diffusion process on the protein structure decoy reduces the similarity between inputs and labels, and when the Structure Refinement step is carried out afterward, the model is unable to avoid learning by memorization of the inputs. This can help alleviate the previously mentioned dilemma of difficult trade-offs, help reduce the mutual information between the learned representations and inputs, increase the mutual information between representations and labels, and finally produce more general and powerful representations.

\subsection{RefineDiff: Refine the Diffused Protein Structure Decoy}
We propose a \emph{novel}, \emph{data-efficient}, and \emph{protein-specific} pretext task for \textbf{3D Geometric Pretraining} named RefineDiff. It consists of two steps, a forward diffusion process for structure decoys, followed by a protein refinement step. The combination of these two steps allows structure decoys to jump out of the local minima of the folding energy landscape and enables the representation model to gain deeper understanding of the protein folding space without prematurely falling into local optima, resulting in generalized representations that are applicable to various downstream tasks. 
\subsubsection{The Forward Diffusion Process}
Instead of reducing the amino acid molecule to a single $C_{\alpha}$ atom, we pursued a finer-grained modeling, so we considered all the amino acid backbone atoms $(C, N, O, C_{\alpha})$. Considering physical plausibility (the bond lengths and bond angles between backbone atoms are relatively fixed, i.e. the relative positions between backbone atoms can be considered fixed), we modeled the backbone atoms as a frame, such that an amino acid backbone structure is determined by two vector properties: translation ${\mathbf{t}}$ (coordinates of the $C_{\alpha}$ atom) and orientation $\mathcal{O}$ (which determines the final coordinates):
\vspace{0.5em}
\begin{equation}
\begin{small}
\mathcal{P} = \{(\mathbf{t}_i, \mathcal{O}_i)\}_{i=i}^{i=N} \ \ \ \text{and} \ \ \ x^a_i = \mathcal{O}x^a_{lit} + \mathbf{t}
\label{equ:5}
\end{small}
\vspace{0.5em}
\end{equation}
where $\mathcal{P}$ is the protein backbone structure, N is the length of the protein sequence, $x^a_i$ is the $a$-type backbone atom of residue i ($a\in {C, N, O, C_{\alpha}}$). $\mathcal{O}x^a_{lit}$ is the coordinate of the $a$-type atom when the $C_{\alpha}$ atom is at the origin and under the unit orthogonal group, which is the normalized coordinate of the amino acid backbone atom\cite{jumper2021highly}.

At this point, to perturb the conformation of the structure decoys, both $\mathbf{t}\in R^3$ and $\mathcal{O}\in SO(3)$ vectors need to be processed. $\mathbf{t}$ is a regular continuous variable that can be perturbed by adding Gaussian noise:
\vspace{0.5em}
\begin{equation}
\begin{small}
\begin{aligned}
q(\mathbf{t}^T_j \ | \ \mathbf{t}^{T-1}_j) = \mathcal{N}(\mathbf{t}^T_j | \sqrt{1-\beta^T_{pos}} \cdot \mathbf{t}^{T-1}_j, \beta^T_{pos}\boldsymbol{I}) ,\\
q(\mathbf{t}^T_j \ | \ \mathbf{t}^0_j) = \mathcal{N}(\mathbf{t}^T_j | \sqrt{\Bar{\alpha}^T_{pos}} \cdot \mathbf{t}^0_j, (1 - \Bar{\alpha}^T_{pos})\boldsymbol{I})
\end{aligned}
\end{small}
\label{equ:6}
\vspace{0.5em}
\end{equation}
where $T$ represents the diffusion time step which is sampled randomly from $[1,T]$ in practice, $\beta^T_{pos}$ means the variance schedule which controls the rate of diffusion and its value increases from 0 to 1 as timp step goes from 0 to t. $\Bar{\alpha}^T_{pos} = \prod^t_{T=1}(1 - \beta^T_{pos})$.

However, $\mathcal{O}$ belongs to the $SO(3)$ group, and we cannot add simple Gaussian noise to it. Therefore we use the method in SE(3) diffusion\cite{leach2022denoising,luo2022antigen} to directly perform the forward Gaussian process for the variable $\mathcal{O}\in SO(3)$. 
\vspace{0.5em}
\begin{equation}
\begin{small}
q(\mathcal{O}^T_j\ |\ \mathcal{O}^0_j) = \mathcal{IG}_{SO(3)}(\mathcal{O}^T_j | \lambda(\sqrt{\Bar{\alpha}^T_{ori}}, \mathcal{O}^0_j), 1 - \Bar{\alpha}^T_{ori})
\label{equ:7}
\end{small}
\vspace{0.5em}
\end{equation}
$IG_{SO(3)}$ denotes the isotropic Gaussian distribution on SO(3) parameterized by a mean rotation and a scalar variance\cite{leach2022denoising}. $\lambda(\gamma, x) = exp(\gamma\log(x))$ is the geodesic flow from $\boldsymbol{I}$ to $x$ by the amount of $\gamma$.

Together, these two processes constitute the forward diffusion process of RefineDiff.

\subsubsection{The Structure Refinement Step}
The goal of the Protein Refinement step is to teach the representation model to comprehend the protein folding space and the folding energy landscape. For this, the model must be able to forecast the direction of conformational optimization, or the direction in which the system will minimize its energy, based on the model's learned understanding of the folding energy landscape and the current conformations of structure decoys that have already undergone the forward diffusion process. 

Given a protein structure decoy $\mathcal{P}$ that has undergone a forward diffusion process, the model needs to predict the direction of conformational change for further optimization, including translation change $\Delta\mathbf{t}\in R^3$ and orientation change $\Delta\mathcal{O}\in SO(3)$ for every residue. We can formulate this process as:
\vspace{0.5em}
\begin{equation}
\begin{small}
\begin{aligned}
\{(\Delta\mathbf{t}_i, \Delta\mathcal{O}_i)\}_{i=1}^{i=N} = \mathcal{E}(\mathcal{P}) \\
\mathbf{t}_i = \mathbf{t}_i + \Delta\mathbf{t}_i\ \text{and}\ \mathcal{O}_i = \mathcal{O} \circ \Delta\mathcal{O}
\label{equ:8}
\end{aligned}
\end{small}
\vspace{0.5em}
\end{equation}
where $\mathcal{E}$ is the equivariant neural network and $\circ$ corresponds to the composition of elements in SO(3) group. We repeat the above process several times for further optimization.

In 3D space, even the same protein structure may differ significantly in coordinates due to global translation or rotation. Therefore structure decoy and ground truth structure need to be aligned with each other first. To avoid the time-consuming process of structural alignment, we use the MSE loss under the local frame\cite{jumper2021highly} as the training objective:
\vspace{0.5em}
\begin{equation}
\begin{small}
\mathcal{L}_{MSE} = \text{Mean}_{i,j}\sqrt{\Vert T^{-1}_i \circ \overrightarrow{x}_{j} - T_i^{true-1} \circ \overrightarrow{x}^{true}_{j} \Vert^2}
\label{equ:9}
\end{small}
\vspace{0.5em}
\end{equation}
where $i\in\{1,\cdots, N_{res}\}, j\in\{C,N,O,C_{\alpha}\}$, $T_i$ and 
$T^{true}_i$ corresponds to the predicted and ground truth frame $(\mathcal{O},\mathbf{t})$ of residue i. And $T_i^{-1} \circ x_j = \mathcal{O}^{-1}(x_j-t_i)$ converts the coordinates of backbone atoms from global to local frame.

Since we predict the frame of each residue separately, the directly connected residues may be too far apart, which is not physically realistic. Therefore, we set up an additional loss function to help the model learn the inter-residue bonds(peptide bonds):
\vspace{0.5em}
\begin{equation}
\begin{small}
\mathcal{L}_{bond} = \frac{1}{N_{bonds}}\sum_{i=1}^{N_{bonds}}\max(\vert \mathbf{l}^i_{pred} - \mathbf{l}^i_{lit}\vert - \mathbf{r}, 0)
\label{equ:10}
\end{small}
\vspace{0.5em}
\end{equation}
where $\mathbf{l}^i_{pred}$ is the bond length in a predicted structure and $\mathbf{l}^i_{lit}$ is the idealized peptide bond length. $N_{bonds}$ is the number of bonds in the structure, and $r$ is the tolerance threshold.
\subsubsection{The Link between RefineDiff and Score Matching}
RFDifussion\cite{watson2022broadly} illustrates the relationship between protein structure refinement and protein structure generation. A good protein structure refinement model can be naturally used as a denoiser in the protein structure DDPM framework, where the generative process can be formulated as follows:
\vspace{0.5em}
\begin{equation}
\begin{small}
p(\mathbf{t}^{T-1}_j|\mathcal{P}^T_j) = \mathcal{N}(\mathbf{t}^{T}_j|\mu_{\theta}(\mathcal{P}^T), \beta^{T-1}_{pos}\boldsymbol{I})
\label{equ:11}
\end{small}
\vspace{0.5em}
\end{equation}
where $\mu_{\theta}(\cdot)$ is a protein refinement model that removes the standard Gaussian noise $\epsilon_j \sim \mathcal{N}(0,\boldsymbol{I})$ added to real structures. The same holds for denoising orientation $\mathcal{O}$.

And at this time their training objectives can be both written in the form of denoising score matching:
\vspace{0.5em}
\begin{equation}
\begin{small}
\mathcal{L}_{pos} = \mathbb{E}[\frac{1}{N_{res}}\sum_j\Vert\epsilon_j - \mathcal{E}(\mathcal{P}^T) \Vert^2].
\label{equ:12}
\end{small}
\vspace{0.5em}
\end{equation}
Here $\mathcal{E}$ is an equivariant neural network that predicts the standard Gaussian noise added to the real structure. As seen above, RefineDiff can be naturally associated with denoising score matching when we perform the forward diffusion process and the PSR step on the real structure.

An important difference between the RefineDiff task and the above scenario is that alignment is required between the structure decoys and the ground truth structure, which leads to the difficulty of calculating score matching. However, we can make a natural generalization of RefineDiff so that it can be performed on the real structure where we replace structure decoys with real structures in the original RefineDiff task. This eliminates the need for structure decoys construction and allows us to obtain both the representation model and the generation model at the same time, like works in 3D point clouds\cite{luo2021diffusion}. We leave this generalization for future work.
\subsection{Model Architecture}
In this subsection, we will briefly describe the neural network architecture used in the pretraining process and in the downstream tasks. As shown in section \ref{framework}, the model architecture includes a 3D geometry-aware representation network as well as 3D prediction heads for 3D geometric pretraining and task-specific projection heads for downstream tasks.

First, we employ Multiple Layer Perceptrons(MLPs) to generate sequential and pairwise embeddings for protein structure decoys. The sequential embedding MLPs map residue features containing information of amino acid type, torsion angle, and 3D coordinates of amino acid backbone atoms to embedding vectors $\{m_i\}_{i=1}^{N_{res}}$. And the pairwise embedding MLPs convert the geometric distance map and dihedral angle map\cite{dauparas2022robust} into feature vectors $\{z_{ij}\}_{i,j=1}^{N_{res}}$. Next, we interactively update the sequential embedding and pairwise embedding then fuse the two into a single sequence representation $\{s_i\}_{i=1}^{N_{res}}$ using shallow layers of sequential attention and triangle attention\cite{lin2022language, jumper2021highly}.
Furthermore, we adopt an orientation-aware roto-translation invariant transformer\cite{luo2022antigen,jumper2021highly} to encode $\{s_i\}$ and the 3D geometric information of current protein decoy structure $\mathcal{P}$ into hidden representations, because the nature of the proteins will not be changed by translation or rotation and protein structure representations should be invariant under the rotation or translation of protein structures. Finally, the representations are fed into corresponding MLPs as 3D prediction heads to predict $SO(3)$ vector and translation vector $\mathbf{t}_{local}$ in local frame. We obtain $\Delta\mathcal{O}$ by converting $SO(3)$ vector to rotation matrix and calculate $\Delta\mathbf{t}$ in the global frame by $\Delta\mathbf{t}=\mathcal{O}_{cur}\mathbf{t}_{local}$. The frame-based update method ensures that the model is \emph{equivariant} to global translation and rotation. 

\section{Experiments}
In this section, we first introduce the experimental setup for both pretraining and downstream tasks. We then show the performance of the pretrained model on protein structure refinement and standard downstream tasks, including Enzyme Commission number prediction, and Gene Ontology term prediction following \cite{gligorijevic2021structure}.
\subsection{Experimental Setup}
\noindent \textbf{Pretraining Dataset.}
We use the DeepAccNet protein structure decoy dataset\cite{hiranuma2021improved} for pretraining. This dataset contains 7992 protein targets (retrieved from the PISCES server and deposited to PDB by May 2018) with limited sequence length (50-300), maximum sequence redundancy of 40\% and minimum resolution of 2.5 \AA. Since our pretraining task has an actual physical scenario, we use the dataset of the CASP14 Protein Refinement track\cite{simpkin2021evaluation} as a test set to examine the performance of the pretrained model on the protein refinement task.

\noindent \textbf{Downstream Tasks.}
We adopt four tasks proposed in \cite{gligorijevic2021structure} as downstream tasks for evaluation. \textbf{\emph{Enzyme Commission (EC) number prediction}} aims to forecast the EC numbers of various proteins, which describe their catalysis of biological activities in a tree structure. It's a multi-label classification task with 538 categories. \textbf{\emph{Gene Ontology(GO) term prediction}} aims to annotate a protein with GO terms which describe the Molecular Function (MF) of gene products, the Biological Processes (BP) in which those actions occur and the Cellular Component (CC) where they are present. Thus, GO term prediction is actually composed of three different sub-tasks: \textbf{\emph{GO-MF, GO-BP, GO-CC}}. 

\noindent \textbf{Baselines and Training.}
We evaluate our model on the benchmark proposed by \cite{zhang2022protein,gligorijevic2021structure} and compare our encoder with various baselines including sequence based encoder( ResNet, LSTM, Transformer\cite{rao2019evaluating}), structure-based encoders( GVP\cite{jing2020learning}, GraphQA\cite{baldassarre2021graphqa}, New IEconv\cite{hermosilla2022contrastive}), large scale pretrained model( ESM-1b\cite{rives2019biological} and GearNet\cite{zhang2022protein}) and co-modeling encoder( DeepFRI\cite{gligorijevic2021structure} and LM-GVP\cite{wang2021lm}). 
Following \cite{gligorijevic2021structure}, we use the multi-cutoff split methods for EC and GO tasks to guarantee the test set contains only PDB chains with sequence identity less than 95\% to the training set. We pretrain our encoder using \emph{RefineDiff} for 150 epochs and then fine-tune it on downstream tasks for 100 epochs. And We train our model on downstream tasks for 100 epochs from scratch. All training is performed on 2 Tesla V100 GPUs.

\noindent \textbf{Evaluation.}
For downstream tasks, we use the protein-centric maximum F-score $F_{max}$ as metric following \cite{zhang2022protein}. Besides, we evaluate performance on PSR with GDT\cite{zemla2003lga} and lDDT\cite{mariani2013lddt}, which are commonly used in CASP\cite{simpkin2021evaluation}. 

\subsection{Performance on Protein Refinement Task}
\begin{table}[!htbp]
\begin{center}
\vspace{-1em}
\caption{Performance on the AlphaFold2 refinement models in CASP14. The performance of our method is highlighted \textbf{boldly}}
\vspace{0.5em}
\label{tab:1}
\resizebox{0.80\columnwidth}{!}{
\scriptsize
\begin{tabular}{c|ccc}
\toprule
\textbf{Method}              & \textbf{GDT-HA}$\uparrow$ & \textbf{GDT-TS} $\uparrow$ & \textbf{lDDT}$\uparrow$ \\ \midrule
Starting            & 70.13                          & 85.56                          & 80.84                        \\ \midrule
GNNRefine           & -3.84                          & -2.14                          & -2.76                        \\
GNNRefine-plus      & -2.69                          & -1.49                          & -2.24                        \\
DeepAccNet          & -5.61                          & -4.69                          & -5.26                        \\
\textbf{RefineDiff} & \textbf{+0.09}                 & \textbf{+0.14}                 & \textbf{+0.12}               \\ \bottomrule
\end{tabular}} \vspace{-1.2em}
\end{center}
\end{table}
Since AlphaFold2 is currently able to achieve relatively high accuracy predictions, the challenge lies in optimizing the results of AlphaFold2. Therefore we evaluate our pretrained model on AlphaFold2 refinement models on CASP14 which is the gold standard in the field of protein structure prediction. From the above table, it can be seen that the structure decoy predicted by AlphaFold2 is deep in the local minima, so after the optimization of other models, the accuracy has decreased instead. However, the model pretrained by RefineDiff can continue to optimize on this basis, although it is not optimized much due to the limitation of model capability. 

This shows that Protein Structure Refinement is learning the protein folding landscape, and when faced with a conformation that is at a local minimum, it is difficult for the model to find further optimization in the critical domain, but instead it pushes the conformation to a worse direction. The forward diffusion process used by RefineDiff helps the conformation escape from the local optimum, and thus allows the model to optimize further.
\vspace{-0.25em}
\subsection{Performance on Downstream Tasks}
\vspace{-0.25em}

We compared with multiple types of baselines, including sequence-based models, structure-based models, co-modeling models and pretrained models on protein sequences as well as on protein structure. Some baselines and tasks are missing due to the computational burden and the lack of codes. We do not include MSA-based baselines since these evolution-based methods require a lot of resources to search and store protein homology sequences following \cite{zhang2022protein}. Here, GearNet is the short for GearNet-Edge-IEConv pretrained with Multi-view Contrastive which is the model with generally best performance in their original paper.
\vspace{-0.5em}

\begin{table}[!htbp]
\begin{center}
\vspace{-1em}
\caption{$F_{max}$ on EC and GO following benchmark propose by \cite{zhang2022protein}, where \textbf{bold} and \underline{underline} denote the best and second metrics. We use our pretext task name to represent pretrained model.}
\vspace{-0.5em}
\label{tab:2}
\resizebox{\columnwidth}{!}{
\begin{tabular}{@{}ccccccc@{}}
\toprule
\multirow{2}{*}{\textbf{Category}}           & \multirow{2}{*}{\textbf{Method}} & \multirow{2}{*}{\textbf{\begin{tabular}[c]{@{}c@{}}Pretraining\\ Dataset(Size)\end{tabular}}} & \multirow{2}{*}{\textbf{EC}}       & \multicolumn{3}{c}{\textbf{GO}}                                                                              \\ \cmidrule(l){5-7} 
                                             &                                  &                                                                                               &                                    & \textbf{BP}                        & \textbf{MF}                        & \textbf{CC}                        \\ \midrule
\multicolumn{7}{c}{\textbf{Train from Scratch}}                                                                                                                                                                                                                                                                                     \\ \midrule
\multirow{3}{*}{\textbf{1D-based}}  & ResNet                           & -                                                                                             & 0.605                              & 0.280                              & 0.405                              & 0.304                              \\
                                             & LSTM                             & -                                                                                             & 0.425                              & 0.225                              & 0.321                              & 0.283                              \\
                                             & Transformer                      & -                                                                                             & 0.238                              & 0.264                              & 0.211                              & 0.405                              \\ \midrule
\multirow{3}{*}{\textbf{2D-based}} & GraphQA                          & -                                                                                             & 0.509                              & 0.308                              & 0.329                              & 0.413                              \\
                                             & New IEConv                       & -                                                                                             & 0.735                              & 0.374                              & 0.544                              & 0.444                              \\
                                             & GearNet              & -                                                                                             & \multicolumn{1}{l}{\textbf{0.810}} & \multicolumn{1}{l}{0.400}          & \multicolumn{1}{l}{0.581}          & \multicolumn{1}{l}{0.430}          \\ \midrule
\multirow{2}{*}{\textbf{3D-based}} & GVP                              & -                                                                                             & 0.489                              & 0.326                              & 0.426                              & 0.420                              \\
                                             & \textbf{IPAFormer(ours)}        & -                                                                                             & \multicolumn{1}{l}{0.783}          & \multicolumn{1}{l}{\textbf{0.410}} & \multicolumn{1}{l}{\textbf{0.583}} & \multicolumn{1}{l}{\textbf{0.455}} \\ \midrule
\multicolumn{7}{c}{\textbf{Pretrain and Finetune}}                                                                                                                                                                                                                                                                                  \\ \midrule
\multirow{2}{*}{\textbf{1D + 2D}}        & DeepFRI                          & Pfam \textbf{(10M)}                                                          & 0.631                              & 0.399                              & 0.465                              & 0.460                              \\
                                             & LM-GVP                           & URf100 \textbf{(216M)}                                                    & 0.664                              & 0.417                              & 0.545                              & {$\underline{0.527}$}                        \\ \midrule
\multirow{2}{*}{\textbf{1D to 2D}}      & ESM-1b                           & UniRef50 \textbf{(24M)}                                                      & 0.864                              & 0.452                              & \textbf{0.657}                     & 0.477                              \\
                                             & GearNet       & AFDB \textbf{(805K)}                                                  & \textbf{0.874}                     & \textbf{0.490}                     & $\underline{0.654}$                              & 0.488                              \\ \midrule
\textbf{3D}                        & \textbf{RefineDiff(ours)}       & \multicolumn{1}{l}{DADB \textbf{(8K)}}                               & \multicolumn{1}{l}{{$\underline{0.865}$}}    & \multicolumn{1}{l}{{$\underline{ 0.455}$}}    & \multicolumn{1}{l}{0.648}          & \multicolumn{1}{l}{\textbf{0.528}} \\ \bottomrule
\end{tabular}} \vspace{-1.5em}
\end{center}
\end{table}
Our pretrained model has a large improvement on multiple downstream tasks compared to models trained from scratch on downstream tasks. In addition, We achieved SOTA on the GO-CC task and second best on the GO-BP and EC tasks, but it is worth noting that we used less than 10,000 protein targets, which is \textbf{1\%} of GearNet which are based on protein 2D geometric pretraining. The above results illustrate that our proposed pretext task and 3D geometric pretraining is \textbf{data-efficient}, powerful and promising. 

\subsection{Ablation Studies}
\label{ablation study}
\begin{table}[!htbp]
\begin{center}
\vspace{-1em}
\caption{Ablation studies on pretraining and downstream task EC. $\Delta GDT = GDT_{refine} - GDT_{prev} $ and $\Delta GDT \in [0,1]$.}
\vspace{0.3em}
\label{tab:3}
\resizebox{0.8\columnwidth}{!}{
\begin{tabular}{@{}lcc@{}}
\toprule
\textbf{Method}                                 & $\Delta GDT$   & $F_{max}$      \\ \midrule
\textbf{IPAFormer}                              & \textbf{0.12} & 0.783          \\
\small{- w/ single layer of triangle attention} & 0.02          & \textbf{0.791} \\ \midrule
\textbf{IPAFormer (RefineDiff)}                 & \textbf{0.12} & \textbf{0.865} \\
\small{- pretrain with PSR directly} & 0.00          & 0.824          \\
\small{- use early checkpoints}                 & 0.01          & 0.847          \\
\small{- w/ single layer of triangle attention} & 0.02          & 0.853          \\ \bottomrule
\end{tabular}} \vspace{-1.2em}
\end{center}
\end{table}

In this section, we compare the model that directly uses the protein structure refinement as the pretext task with the model that uses RefineDiff for pretraining. From the table above, we can see that it is difficult for the model to further optimize the protein structure decoy when trained directly using protein structure refinement, and the improvement in performance is limited when migrating to downstream tasks compared to pretraining with RefineDiff.

Interestingly, despite the fact that early in the training of RefineDiff the capability of pretrained model for the protein structure decoy refinement is not that strong , the model using RefineDiff as the pretraining task still achieves better performance on the downstream task than the model trained directly using PSR as the pretraining task. This suggests that even if the model does not learn RefineDiff well due to insufficient training or ability, it can still benefit from seeing a more complete energy landscape compared to using PSR directly as a pretraining task.

\section{Conclusion}
\vspace{-0.3em}
In this paper, we present the development of protein pretraining and propose a unified framework; then we analyze the current problems of 2D protein structure pretraining (data homogeneity, physically unrealistic modeling and limited pretext task), and point out the need for 3D geometric pretraining on proteins. We take the pretext task as a starting point and propose a \emph{data-efficient} and \emph{protein-specific} pretext task for \emph{3D geometric pretraining on proteins}. After pretraining our geometric-aware model with this task on limited data(less than 1\% of SOTA models), we obtained informative protein representations that can achieve comparable performance for various downstream tasks. 

\bibliography{example_paper}
\bibliographystyle{icml2022}

\newpage
\appendix
\onecolumn
\section{Appendix.}

\subsection{More Related Work}
\noindent \textbf{Protein Model Quality Assessment.}
Some work uses neural networks to predict the accuracy of Decoy (which is a twin problem of PSR called Model Quality Assessment, MQA) as a way to guide conformational sampling toward the lowest point of folding energy\cite{hiranuma2021improved}. Other works model at the pseudo-3D level, starting from the scalar feature Distance Map of the protein, optimizing the prediction of the Distance Map by a neural network such as GNN, and then using tools such as Rosetta to perform conformational sampling based on the optimized Distance Map\cite{jing2021fast}.

\noindent \textbf{Protein Representation Model.}
Protein representation models are one of the core challenges in applying deep learning to protein understanding, and they evolve as deep learning does\cite{wu2022survey}. Initially, researchers attempted to represent proteins using convolutional neural networks, which were typically based on handcrafted feature maps\cite{yang2022convolutions,derevyanko2018deep,amidi2018enzynet}; however, with the advancement of Natural Language Processing(NLP) and the similarities between amino acid sequences and natural language, powerful NLP models such as Transformer were introduced to protein representation\cite{rao2020transformer,rao2021msa,lin2022language}. Simultaneously, graph neural networks were applied to protein structure representation\cite{baldassarre2021graphqa,gligorijevic2021structure}, with promising results in fields such as protein structure design\cite{ingraham2019generative,dauparas2022robust}. With the rise of geometric deep learning, Equivariant Neural Networks(ENN) began to be applied to protein representation, allowing us to directly learn protein structures end-to-end and achieve more powerful results\cite{hsu2022learning,jing2020learning,wu2022atomic}. AlphaFold2 is likely the pinnacle of these techniques, revolutionizing structural biology and becoming one of the best instances of AI applications. Inspired by this, many works have tried to combine sequence and structure representation, expecting that co-modeling can combine the advantages of both modalities.\cite{lin2022language,mansoor2021toward,you2022cross}

\subsection{More Analysis}
Since the current structure pretraining does not directly model 3D structural data, the structural features involved are scalar, so 2D Graph-based GNN can be used, which is why we call it pseudo-3D geometric pretraining. There have been few attempts in 3D geometric pretraining.\cite{mansoor2021toward}

Compared with the sequence pretraining model, the 2D geometry pretraining model handles more complex data, and its training efficiency is reduced by orders of magnitude\cite{zhang2022protein}. However, by constructing structural scalar features and using structure-related scalar prediction tasks, it successfully achieves geometry awareness, so it can achieve comparable results with sequence pretraining with much less data (1\% of sequence pretraining), i.e., data-efficient\cite{mansoor2021toward}.

\end{document}